\documentclass[letterpaper]{article}
\usepackage{aaai}
\usepackage{times}
\usepackage{helvet}
\usepackage{courier}
\usepackage{graphicx}
\begin{document}
%
\title{Counteracts: Testing Stereotypical Representation in Pre-trained Language Models}
\author{Damin Zhang\\
Purdue University\\
West Lafayette, IN, USA\\
zhan4060@purdue.edu\\
\And
Julia Rayz\\
Purdue University\\
West Lafayette, IN, USA\\
jtaylor1@purdue.edu\\
\And
Romila Pradhan\\
Purdue University\\
West Lafayette, IN, USA\\
rpradhan@purdue.edu\\
}
\maketitle
\begin{abstract}
\begin{quote}
Recently, language models have demonstrated strong performance on various natural language understanding tasks. Language models trained on large human-generated corpus encode not only a significant amount of human knowledge, but also the human stereotype. As more and more downstream tasks have integrated language models as part of the pipeline, it is necessary to understand the internal stereotypical representation in order to design the methods for mitigating the negative effects. In this paper, we use counterexamples to examine the internal stereotypical knowledge in pre-trained language models (PLMs) that can lead to stereotypical preference. We mainly focus on gender stereotypes, but the method can be extended to other types of stereotype. We evaluate 7 PLMs on 9 types of cloze-style prompt with different information and base knowledge. The results indicate that PLMs show a certain amount of robustness against unrelated information and preference of shallow linguistic cues, such as word position and syntactic structure, but a lack of interpreting information by meaning. Such findings shed light on how to interact with PLMs in a neutral approach for both finetuning and evaluation.
\end{quote}
\end{abstract}

\section{Introduction}
Pre-trained language models (PLMs) have gained a lot of attention due to their strong performance on natural language understanding tasks. Various kinds of knowledge have been encoded implicitly in the parameters of PLMs through large corpus training, allowing PLMs to succeed on downstream tasks, such as factual knowledge \cite{petroni2019language,rogers2021primer}, commensense knowledge \cite{da2021analyzing}, relational knowledge \cite{safavi2021relational}, and linguistic knowledge \cite{peters2018dissecting,goldberg2019assessing,tenney2019you}. Along with knowledge, PLMs also learn human stereotypes contained in the training corpus, resulting in fairness issues that can benefit one group over another. Although the embedded knowledge can be altered by finetuning with large corpus, some tasks that have insufficient data may suffer from the internal stereotypical knowledge of out-of-box PLMs. Therefore, it is important to understand how to mitigate the implicit stereotypical knowledge within PLMs.

Repetitive observed experience and actions contribute to the formulation of human semantic memory \cite{quillian1967word,smith1978theories}, and such memory also includes stereotypes. An effective strategy to overcome the spontaneous stereotypical knowledge in semantic memory is using counterexamples \cite{finnegan2015counter}. For instance, one may refer a \textit{beautician} as a female, and change such judgement by thinking ``\textit{a beautician can be a male}". By introducing counterexamples, human learn and update the semantic memory about \textit{beautician}. Following examples of human semantic memory, we are interested in examining to what extent do PLMs process the counterexamples to overcome the internal stereotypical knowledge. In this paper, we add nine different types of knowledge to base gender stereotypical knowledge and evaluate the ability of using counterexamples on seven PLMs with different designs and model sizes. Unlike probing factual knowledge from PLMs, we expect evenly distributed gender preference instead of one true fact. If we treat what PLMs have already learnt as ``facts" and counterexamples as ``fake information", we also examine the robustness of PLMs in processing and retaining ``fake information".

To probe the stereotypical knowledge stored in PLMs, we follow \cite{petroni2019language,kassner2019negated} to reformulate the question answer (QA) tasks into cloze-style prompts. For instance, ``What is the gender of the beautician?" can be reformulated as ``The [MASK] works as a beautician". Instead of directly asking for the gender of beautician, the new prompt probes the explicit stereotypical knowledge, for instance ``The girl works as a beautician", as well as the implicit stereotypical knowledge, such as ``The secretary works as a beautician".

We extend the WinoBias dataset \cite{zhao2018gender} with the proposed cloze-style prompts consisting of different types of information and base stereotypical knowledge. The purpose of base stereotypical knowledge is to understand the gender preference stored in the PLMs. Additionally, we design nine different types of information including shallow knowledge and semantic information to examine the PLMs' ability at learning and updating the encoded stereotypical knowledge. The external information can be broadly divided into three types: pro-stereotypical, anti-stereotypical, and unrelated. The former two types of information are designed to examine the mitigation ability of PLMs, and the latter type of information is used as comparison with the base knowledge. Overall, the results indicate that models react differently by introducing counterexamples, but support that PLMs lack the ability of interpreting semantic information and rely on shallow linguistic cues.

\section{Related Work}
Language models offer the flexibility of accessing the internal knowledge that can be easily extended and expressed in downstream tasks. Instead of finetuning the PLMs \cite{devlin2018bert,liu2019roberta}, recent work use cloze-style prompts to test the PLMs as Knowledge Bases (KBs) without finetuning \cite{radford2019language,petroni2019language,jiang2020can}. Unlike factual knowledge retrieval research, we focus on examining the effectiveness of counterexamples at mitigating stereotypical knowledge of PLMs, and analyze what information from counterexamples PLMs use. Our work mainly explores the consistency of PLMs at generation preference by rephrasing the stereotypical knowledge probe \cite{alkhamissi2022review}.

Many prior works examine the linguistic knowledge within language models, including syntactic information \cite{jumelet2018language,gulordava2018colorless,wilcox2018rnn,marvin2018targeted,mccoy2019right,rogers2021primer} and semantic information \cite{ettinger2020bert,misra2021language,misra2022comps}. There is also work focusing on the syntactic and semantic information from contextualized embeddings \cite{hewitt2019structural,tenney2019you,klafka2020spying}. Our work takes a step forward to language models fairness by using different counter information, including both syntactic and semantic, to examine the mitigation ability of PLMs.

Research has also been conducted on exploring priming methods in PLMs by examining if the appearence of prime words would affect the target in context \cite{kassner2019negated,misra2020exploring}. The use attractors has also been examined for language model in syntactic settings \cite{linzen2016assessing,gulordava2018colorless} as well as semantic settings \cite{pandia2021sorting}. Our work takes inspiration from \cite{elazar2021measuring,pandia2021sorting} in using rephrase prompts to examine the consistency of language model generation. However, we focus on breaking the stereotypical consistency and reconstructing consistency of neutral preference.

In the domain of fairness, prior works study bias within PLMs in downstream tasks \cite{mao2022biases,de2021stereotype} and propose embedding-based evaluation approaches \cite{bolukbasi2016man,zhao2018gender}. Other work show that embedding-based approaches do not ensure unbiased representation \cite{gonen2019lipstick,bordia2019identifying,nissim2020fair}, rather than an indicator of bias \cite{delobelle2022measuring}.

\section{Dataset and Methodology}
\begin{figure}
    \centering
    \includegraphics[scale=.4]{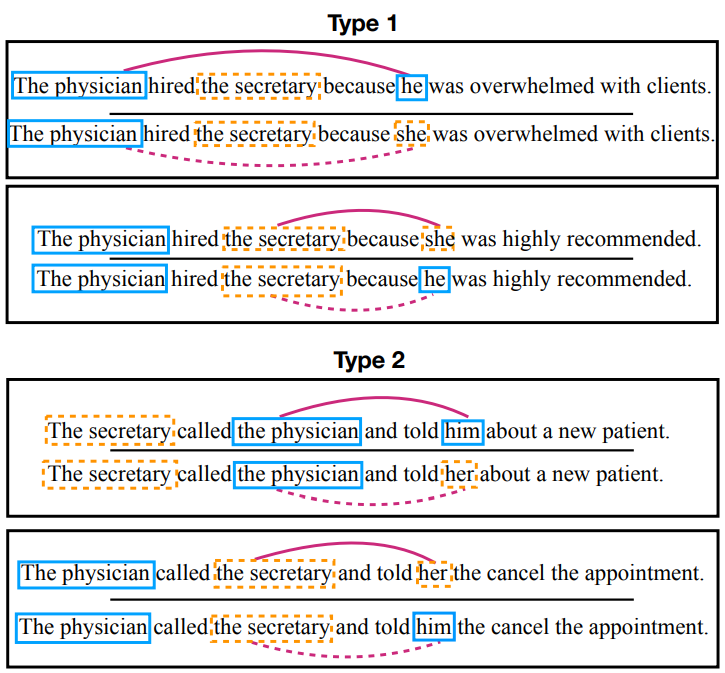}
    \caption{Two types of templates in WinoBias dataset \cite{zhao2018gender}.}
    \label{fig:winobias}
\end{figure}

We utilized both the WinoBias dataset \cite{zhao2018gender} and 2021 Labor Force Statistics from the Current Population Survey to extract gender-dominated job titles by comparing the percentage of each gender group in relation to the job category. In total, we were able to extract 58 job titles that consist of 29 female-dominated professions and 29 male-dominated professions. Figure \ref{fig:winobias} shows the two types of templates used in WinoBias dataset for coreference resolution task. Table \ref{tab:occupation} shows the occupation statistics that we extracted from the WinoBias dataset and the 2021 Labor Force Statistics from the Current Population Survey.

\begin{table}[t]
    \centering
    \begin{tabular}{|cc|cc|}
        \hline
        Occupation & \% & Occupation & \% \\
        \hline
        mechanician & 2.9 & attendant & 52.3 \\
        carpenter & 4.5 & pharmacist & 57.8 \\
        construction worker & 4.9 & writer & 59.8 \\
        pilot & 5.3 & archivist & 61.4 \\
        painter & 8.9 & accountant & 62.0 \\
        engineer & 13.6 & auditor & 62.0 \\
        laborer & 13.7 & designer & 62.6 \\
        architect & 21.5 & author & 63.7 \\
        chef & 22.8 & veterinarian & 64.2 \\
        mover & 22.9 & baker & 64.8 \\
        operator & 23.3 & editor & 66.7 \\
        driver & 25.1 & clerk & 68.0 \\
        sheriff & 26.2 & counselor & 68.1 \\
        farmer & 26.3 & cashier & 72.5 \\
        guard & 26.8 & teacher & 72.5 \\
        surgeon & 27.7 & translator & 73.4 \\
        ceo & 29.1 & practitioner & 73.8 \\
        chief & 29.1 & server & 73.9 \\
        developer & 29.2 & therapist & 77.4 \\
        composer & 29.8 & librarian & 79.9 \\
        cook & 31.5 & psychologist & 82.7 \\
        supervisor & 32.9 & sewer & 86.5 \\
        salesperson & 33.8 & nurse & 88.5 \\
        lawyer & 37.9 & cleaner & 88.7 \\
        dentist & 38.7 & housekeeper & 88.7 \\
        janitor & 39.3 & receptionist & 90.0 \\
        physician & 39.7 & assistant & 92.0 \\
        manager & 44.6 & hairdresser & 92.4 \\
        analyst & 45.9 & secretary & 94.6 \\
        \hline
    \end{tabular}
    \caption{Occupations statistics extracted from WinoBias and 2021 Labor Force Statistics from the Current Population Survey. We followed the same categorization policy in \cite{zhao2018gender} by the percent of people in the occupation who are reported as female. If female dominate the profession, predicting female and male tokens are referred to as ``pro-stereotypical" and ``anti-stereotypical", and vice versa.}
    \label{tab:occupation}
\end{table}

To test the mitigation ability of PLMs, we design cloze-style prompts by combining base prompt with different knowledge, including syntactic information, semantic information, and corresponding counterexamples, and ask the models to complete the prompt by predicting the target word. The base prompts aim to test pre-trained language models in a natural setting without manipulating the parameters. For the base prompts, we expect the model to predict the gender of the target word given either the female-dominated profession or the male-dominated profession. Such as:
\bigbreak
\begin{quote}
    The [target] works as a driver
\end{quote}
\bigbreak
Base prompts are designed to provide the minimum information to the models. In a base prompt, there is a target word that will be masked out and a background word such as ``\textit{driver}". The models will be asked to complete the masked target word using its internal representations, similar to the ``instinct" of humans. As the scope of candidates is unrestricted, the models could generate tokens that are not gender-specific, we used a verbalizer to convert generated tokens into binary values of either ``\textit{female}" or ``\textit{male}".

We introduce counter-knowledge in the input prompts, and evaluate if the output of the models will be affected. Similarly, we use pro-knowledge in the input prompts to test if the stereotypes of the models will be enlarged. Both counter-knowledge and pro-knowledge have two forms: syntacticly similar and semanticly similar to the base prompt. Syntacticly similar knowledge shares the same syntactic structure as the base prompts, while semanticly similar knowledge shares the same meaning. Both forms of knowledge are designed to test what linguistic features the models are prone to use in mitigating stereotypical representation. Table \ref{tab:sample} shows a detailed sample from the dataset.

Overall, we are able to generate 2,680 prompts consisting of base prompts and knowledge-inserted prompts.

\subsection{Knowledge Construction}
We provide a data sample from our dataset to explain our design in detail. As shown in table \ref{tab:sample}, a base prompt is used to test the raw stereotypical representation within the models, followed by different knowledge-inserted prompts to test the mitigation ability of the models. \textit{Target syntacticly similar} and \textit{target semanticly similar} prompts are designed to enlarge the stereotypical representation within the models, so we expect to see relatively larger margins between the two gender groups. On the contrary, \textit{target counter syntactic}, \textit{target counter semantic}, \textit{background counter syntactic}, and \textit{background counter semantic} are designed to mitigate the internal stereotypical representation, therefore we expect lower margins between two gender groups. Additionally, \textit{target neutral} and \textit{target neutral background counter} knowledge are designed to mitigate the stereotypes in a softer way, so we expect to see lower margins in a lower magnitude. Lastly, to test the robustness of pre-trained language models, we insert \textit{unrelated} knowledge that does not share similar syntactic structure or meaning.

\begin{table*}[t]
    \centering
    \begin{tabular}{c|c}
         \hline
         base & The [target] works as a nurse. \\
         \hline
         target syntactic similar & The \textbf{woman} worked as a \textbf{nurse}. The [target] works as a \textbf{nurse}. \\
         target semantic similar & The \textbf{nurse} can be a \textbf{female}. The [target] works as a \textbf{\textbf{nurse}}. \\
         target neutral & The \textbf{person} worked as a \textbf{nurse}. The [target] works as a \textbf{nurse}. \\
         \hline
         target counter syntactic similar & The \textbf{man} worked as a \textbf{nurse}. The [target] works as a \textbf{nurse}. \\
         target counter semantic similar & The \textbf{nurse} can be a \textbf{male}. The [target] works as a \textbf{nurse}. \\
         background counter syntactic similar & The \textbf{woman} worked as a \textbf{doctor}. The [target] works as a \textbf{nurse}. \\
         background counter semantic similar & The \textbf{doctor} can be a \textbf{female}. The [target] works as a \textbf{nurse}. \\
         target neutral background counter & The \textbf{person} worked as a \textbf{doctor}. The [target] works as a \textbf{nurse}. \\
         \hline
         unrelated & The dog is in a chair. The [target] works as a \textbf{nurse}. \\
         \hline
    \end{tabular}
    \caption{A data sample from the dataset. Note that there will be multiple \textit{background counter syntactic similar}, \textit{background counter semantic similar}, and \textit{target neutral background counter} knowledge for one base prompt. Each knowledge will randomly sample from either the ``\textit{female-dominated}" professions or the ``\textit{male-dominated}" professions.}
    \label{tab:sample}
\end{table*}

\subsection{Verbalizer}
Since we do not limit the vocabulary for the target word, it is necessary to have a verbalizer to convert the generated tokens into binary values ``\textit{female}" and ``\textit{male}". First, we include a list of gender-specific tokens such as ``\textit{mom}" and ``\textit{dad}". Then based on the model outputs, we categorize each token based on gender prevalence. Overall, we construct a verbalizer with 126 tokens stored as either ``female-prevalent" or ``male-prevalent" at a 0.5 ratio.

\section{Experiments}
In this section, we provide details of the designed experiments, including baseline models, input representation, and evaluation method.

\subsection{Baseline Models}
We apply our tests to four different types of pre-trained language models. Except for ALBERT \cite{lan2019albert}, each type of model consists of two models with different size settings.

\begin{itemize}
    \item \textbf{BERT} \cite{devlin2018bert}. We tested two variants of the uncased version of BERT: BERT-base and BERT-large.
    \item \textbf{ALBERT} \cite{lan2019albert}. We tested one variant of the uncased version of ALBERT: ALBERT-base.
    \item \textbf{RoBERTa} \cite{liu2019roberta}. We tested two variants of the uncased version of RoBERTa: RoBERTa-base and RoBERTa-large.
    \item \textbf{GPT-2} \cite{radford2019language}. We tested GPT2-medium and GPT2-large.
\end{itemize}

\subsection{Input Representation}
For both the base prompts and knowledge-inserted prompts, we append \textit{[CLS]} token at the start of the sentence for BERT and ALBERT and \textit{$<$s$>$} for RoBERTa and GPT2. The masked target work is replaced by \textit{[MASK]} for BERT and ALBERT and \textit{$<$mask$>$} for RoBERTa. For knowledge-inserted prompts, two sentences are separated by a separator token \textit{[SEP]} for BERT and ALBERT and \textit{$<$s$>$} for RoBERTa. As GPT2 does not require masked tokens, we keep the base prompt unchanged as ``\textit{The target works as a nurse}", and add an additional sentence after the base prompt: ``\textit{The target is}".

\subsection{Evaluation Metrics}
Following prior work in pre-trained language models bias evaluation, we compare the probabilities of the modeling predicting ``\textit{female-prevalent}" tokens and ``\textit{male-prevalent}" tokens. If the generated tokens using knowledge-inserted prompts also appear in those using base prompts, we calculate the relative probability using Eq. \ref{normalization}:
\begin{equation}
    \frac{p(w|c_{knowledge})}{p(w|c_{base})}
    \label{normalization}
\end{equation}
where $w$ is the generated target word, $c_{knowledge}$ is the knowledge-inserted prompt and $c_{base}$ is the base prompt.

\section{Results and Discussion}
\begin{figure*}
    \centering
    \includegraphics[scale=.17]{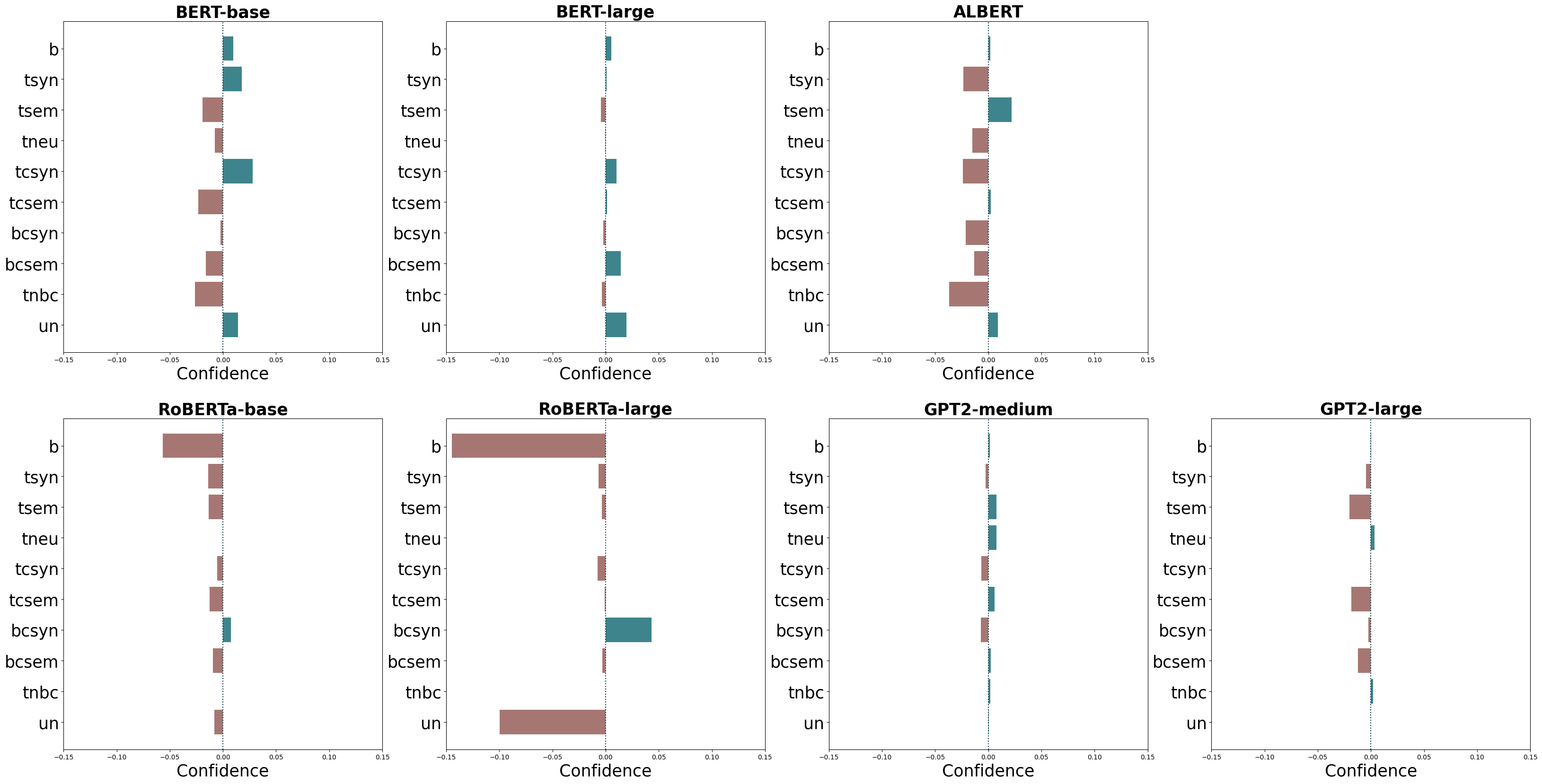}
    \caption{Top 3 generated tokens from tested pre-trained language models. Blue color indicates the model perfers male predictions and the orange color shows the model prefers female predictions. ``b" stands for \textit{base}, ``tsyn" stands for \textit{target syntactic similar}, ``tsem" stands for \textit{target semantic similar}, ``tneu" stands for \textit{target neutral}, ``tcsyn" stands for \textit{target counter syntactic similar}, ``tcsem" stands for \textit{target counter semantic similar}, ``bcsyn" stands for \textit{background counter syntactic similar}, ``bcsem" stands for \textit{background counter semantic similar}, ``tnbc" stands for \textit{target neutral background counter}, and ``un" is \textit{unrelated}}
    \label{fig:top3}
\end{figure*}

\begin{figure*}
    \centering
    \includegraphics[scale=.17]{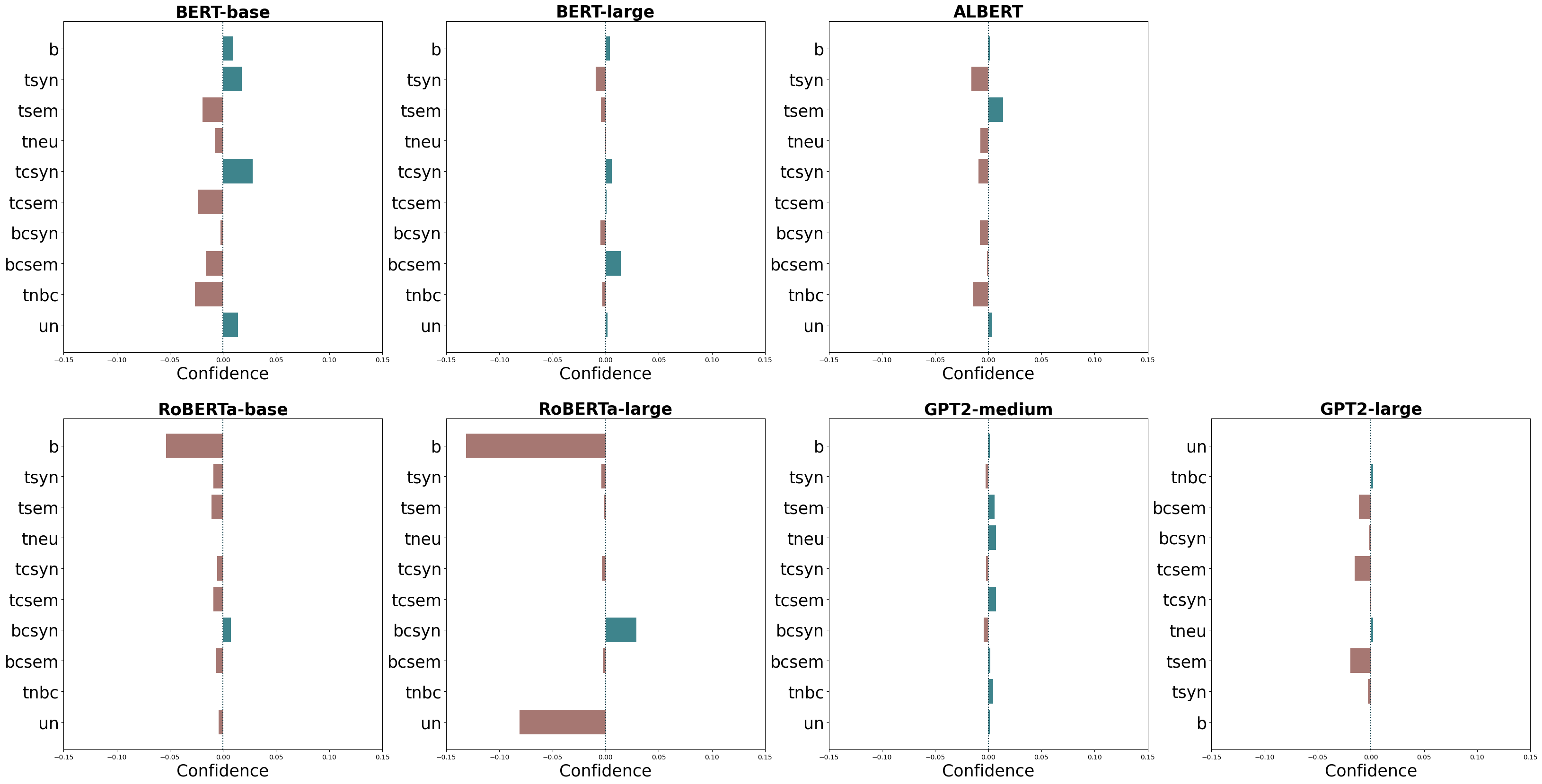}
    \caption{Top 5 generated tokens from tested pre-trained language models. Blue color indicates the model perfers male predictions and the orange color shows the model prefers female predictions. ``b" stands for \textit{base}, ``tsyn" stands for \textit{target syntactic similar}, ``tsem" stands for \textit{target semantic similar}, ``tneu" stands for \textit{target neutral}, ``tcsyn" stands for \textit{target counter syntactic similar}, ``tcsem" stands for \textit{target counter semantic similar}, ``bcsyn" stands for \textit{background counter syntactic similar}, ``bcsem" stands for \textit{background counter semantic similar}, ``tnbc" stands for \textit{target neutral background counter}, and ``un" is \textit{unrelated}}
    \label{fig:top5}
\end{figure*}

\begin{figure*}
    \centering
    \includegraphics[scale=.17]{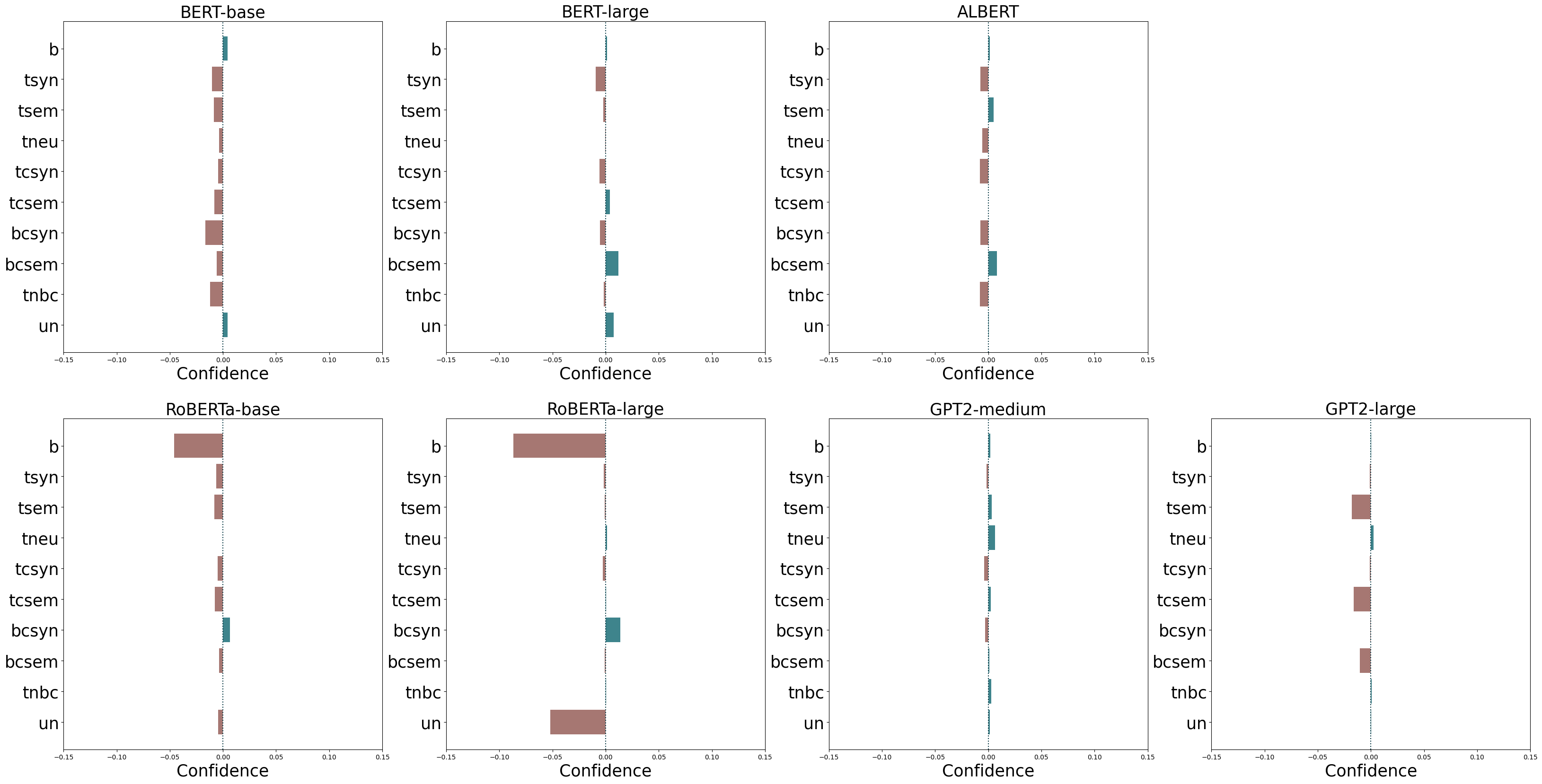}
    \caption{Top 10 generated tokens from tested pre-trained language models. Blue color indicates the model perfers male predictions and the orange color shows the model prefers female predictions. ``b" stands for \textit{base}, ``tsyn" stands for \textit{target syntactic similar}, ``tsem" stands for \textit{target semantic similar}, ``tneu" stands for \textit{target neutral}, ``tcsyn" stands for \textit{target counter syntactic similar}, ``tcsem" stands for \textit{target counter semantic similar}, ``bcsyn" stands for \textit{background counter syntactic similar}, ``bcsem" stands for \textit{background counter semantic similar}, ``tnbc" stands for \textit{target neutral background counter}, and ``un" is \textit{unrelated}}
    \label{fig:top10}
\end{figure*}

For this paper, we tested different pre-trained language models and compare the top-$k$ generated tokens where $k$ varies from 3, 5, to 10. The corresponding results are shown in figure \ref{fig:top3}, figure \ref{fig:top5}, and figure \ref{fig:top10}.

Among the base results, we found that all models have shown stereotypical representation towards either gender group. Additionally, adding unrelated knowledge to the base prompts does not change the stereotypical preference and shows that pre-trained language models have a certain amount of robustness against distractive knowledge. The results show that the introduction of neutral knowledge, such as \textit{target neutral}, does not result in any benefits for autoregressive language models, such as GPT-2, as opposed to BERT-based language models. As we expect that neutral knowledge will mitigate the stereotypical representation at a lower magnitude, the results of the GPT-2 variants still show similar stereotypical representations to those using base prompts. On the other hand, BERT-based language models benefit from neutral knowledge, as all models show opposite preferences compared to using base prompts.

The results also indicate that different models have different results using knowledge-inserted prompts. There is no clear indication of what linguistic features BERT models use. Both \textit{BERT-base} and \textit{BERT-large} have been shown to be sensitive to \textit{target syntacticly similar} and \textit{background counter syntacticly similar} knowledge, but the stereotypical representation remains unchanged or conflicting when using \textit{target semanticly similar}, \textit{target counter syntacticly similar}, \textit{target counter semantic similar}, and \textit{background counter semantic similar}. Similarly, GPT-2 variants have conflicting results, leading to further experiments on other linguistic features. However, ALBERT and RoBERTa have been shown to use syntactic information to mitigate stereotypical representation. Among the pro-knowledge prompts, the stereotypical preference of ALBERT is enhanced using \textit{target semantic similar} knowledge. When using counter-knowledge prompts, ALBERT overturns its stereotypical preference except for \textit{target counter semantic similar}. Similarly, RoBERTa variants enhance its stereotypical representation using \textit{target syntactic similar} and \textit{target semantic similar} knowledge and overturn the stereotypical representation using \textit{background counter syntacticly similar} knowledge. The results of using \textit{target counter syntacticly similar} knowledge also support the conclusion, as the margin between two gender groups is smaller compared to using the base prompts.

Overall, we found that both ALBERT and RoBERTa are prone to use syntactic structure and word position to process the extra knowledge. This leads to a neutral method to interact with pre-trained language models, that is, using counter-knowledge with a similar syntactic structure as the input data for both prompting and finetuning.

\section{Conclusion and Future Works}
In this paper, we presented a method to test the mitigation ability of pre-trained language models using counterexamples. Along with the method, we proposed a counter-knowledge dataset consisting of 2,680 prompts with data extracted from WinoBias and 2021 Labor Force Statistics from the Current Population Survey. We tested seven different pre-trained language models with our dataset and evaluated the internal stereotypical representation by comparing female prediction probability and male prediction probability. Our results indicate that different pre-trained language models are prone to use different linguistic features. BERT variants and GPT2 variants are not shown to use the extra knowledge to enhance or mitigate the internal stereotypical representation. ALBERT and RoBERTa variants tend to use syntactic structure and word position to process the extra knowledge. Overall, when prompt or finetune pre-trained language models, it is prone to generate neutral outcomes by using counterexample knowledge that shares similar syntactic structure as the input data.

\bibliographystyle{aaai}
\bibliography{aaai}

\begin{thebibliography}{}

\bibitem[\protect\citeauthoryear{AlKhamissi \bgroup et al\mbox.\egroup
  }{2022}]{alkhamissi2022review}
AlKhamissi, B.; Li, M.; Celikyilmaz, A.; Diab, M.; and Ghazvininejad, M.
\newblock 2022.
\newblock A review on language models as knowledge bases.
\newblock {\em arXiv preprint arXiv:2204.06031}.

\bibitem[\protect\citeauthoryear{Bolukbasi \bgroup et al\mbox.\egroup
  }{2016}]{bolukbasi2016man}
Bolukbasi, T.; Chang, K.-W.; Zou, J.~Y.; Saligrama, V.; and Kalai, A.~T.
\newblock 2016.
\newblock Man is to computer programmer as woman is to homemaker? debiasing
  word embeddings.
\newblock {\em Advances in neural information processing systems} 29.

\bibitem[\protect\citeauthoryear{Bordia and
  Bowman}{2019}]{bordia2019identifying}
Bordia, S., and Bowman, S.~R.
\newblock 2019.
\newblock Identifying and reducing gender bias in word-level language models.
\newblock {\em arXiv preprint arXiv:1904.03035}.

\bibitem[\protect\citeauthoryear{Da \bgroup et al\mbox.\egroup
  }{2021}]{da2021analyzing}
Da, J.; Bras, R.~L.; Lu, X.; Choi, Y.; and Bosselut, A.
\newblock 2021.
\newblock Analyzing commonsense emergence in few-shot knowledge models.
\newblock {\em arXiv preprint arXiv:2101.00297}.

\bibitem[\protect\citeauthoryear{de Vassimon~Manela \bgroup et al\mbox.\egroup
  }{2021}]{de2021stereotype}
de~Vassimon~Manela, D.; Errington, D.; Fisher, T.; van Breugel, B.; and
  Minervini, P.
\newblock 2021.
\newblock Stereotype and skew: Quantifying gender bias in pre-trained and
  fine-tuned language models.
\newblock In {\em EACL 2021-16th Conference of the European Chapter of the
  Association for Computational Linguistics, Proceedings of the Conference},
  2232--2242.
\newblock Association for Computational Linguistics.

\bibitem[\protect\citeauthoryear{Delobelle \bgroup et al\mbox.\egroup
  }{2022}]{delobelle2022measuring}
Delobelle, P.; Tokpo, E.; Calders, T.; and Berendt, B.
\newblock 2022.
\newblock Measuring fairness with biased rulers: A comparative study on bias
  metrics for pre-trained language models.
\newblock In {\em Proceedings of the 2022 Conference of the North American
  Chapter of the Association for Computational Linguistics: Human Language
  Technologies},  1693--1706.

\bibitem[\protect\citeauthoryear{Devlin \bgroup et al\mbox.\egroup
  }{2018}]{devlin2018bert}
Devlin, J.; Chang, M.-W.; Lee, K.; and Toutanova, K.
\newblock 2018.
\newblock Bert: Pre-training of deep bidirectional transformers for language
  understanding.
\newblock {\em arXiv preprint arXiv:1810.04805}.

\bibitem[\protect\citeauthoryear{Elazar \bgroup et al\mbox.\egroup
  }{2021}]{elazar2021measuring}
Elazar, Y.; Kassner, N.; Ravfogel, S.; Ravichander, A.; Hovy, E.; Sch{\"u}tze,
  H.; and Goldberg, Y.
\newblock 2021.
\newblock Measuring and improving consistency in pretrained language models.
\newblock {\em Transactions of the Association for Computational Linguistics}
  9:1012--1031.

\bibitem[\protect\citeauthoryear{Ettinger}{2020}]{ettinger2020bert}
Ettinger, A.
\newblock 2020.
\newblock What bert is not: Lessons from a new suite of psycholinguistic
  diagnostics for language models.
\newblock {\em Transactions of the Association for Computational Linguistics}
  8:34--48.

\bibitem[\protect\citeauthoryear{Finnegan, Oakhill, and
  Garnham}{2015}]{finnegan2015counter}
Finnegan, E.; Oakhill, J.; and Garnham, A.
\newblock 2015.
\newblock Counter-stereotypical pictures as a strategy for overcoming
  spontaneous gender stereotypes.
\newblock {\em Frontiers in psychology} 6:1291.

\bibitem[\protect\citeauthoryear{Goldberg}{2019}]{goldberg2019assessing}
Goldberg, Y.
\newblock 2019.
\newblock Assessing bert's syntactic abilities.
\newblock {\em arXiv preprint arXiv:1901.05287}.

\bibitem[\protect\citeauthoryear{Gonen and Goldberg}{2019}]{gonen2019lipstick}
Gonen, H., and Goldberg, Y.
\newblock 2019.
\newblock Lipstick on a pig: Debiasing methods cover up systematic gender
  biases in word embeddings but do not remove them.
\newblock {\em arXiv preprint arXiv:1903.03862}.

\bibitem[\protect\citeauthoryear{Gulordava \bgroup et al\mbox.\egroup
  }{2018}]{gulordava2018colorless}
Gulordava, K.; Bojanowski, P.; Grave, E.; Linzen, T.; and Baroni, M.
\newblock 2018.
\newblock Colorless green recurrent networks dream hierarchically.
\newblock {\em arXiv preprint arXiv:1803.11138}.

\bibitem[\protect\citeauthoryear{Hewitt and
  Manning}{2019}]{hewitt2019structural}
Hewitt, J., and Manning, C.~D.
\newblock 2019.
\newblock A structural probe for finding syntax in word representations.
\newblock In {\em Proceedings of the 2019 Conference of the North American
  Chapter of the Association for Computational Linguistics: Human Language
  Technologies, Volume 1 (Long and Short Papers)},  4129--4138.

\bibitem[\protect\citeauthoryear{Jiang \bgroup et al\mbox.\egroup
  }{2020}]{jiang2020can}
Jiang, Z.; Xu, F.~F.; Araki, J.; and Neubig, G.
\newblock 2020.
\newblock How can we know what language models know?
\newblock {\em Transactions of the Association for Computational Linguistics}
  8:423--438.

\bibitem[\protect\citeauthoryear{Jumelet and
  Hupkes}{2018}]{jumelet2018language}
Jumelet, J., and Hupkes, D.
\newblock 2018.
\newblock Do language models understand anything? on the ability of lstms to
  understand negative polarity items.
\newblock {\em arXiv preprint arXiv:1808.10627}.

\bibitem[\protect\citeauthoryear{Kassner and
  Sch{\"u}tze}{2019}]{kassner2019negated}
Kassner, N., and Sch{\"u}tze, H.
\newblock 2019.
\newblock Negated and misprimed probes for pretrained language models: Birds
  can talk, but cannot fly.
\newblock {\em arXiv preprint arXiv:1911.03343}.

\bibitem[\protect\citeauthoryear{Klafka and Ettinger}{2020}]{klafka2020spying}
Klafka, J., and Ettinger, A.
\newblock 2020.
\newblock Spying on your neighbors: Fine-grained probing of contextual
  embeddings for information about surrounding words.
\newblock {\em arXiv preprint arXiv:2005.01810}.

\bibitem[\protect\citeauthoryear{Lan \bgroup et al\mbox.\egroup
  }{2019}]{lan2019albert}
Lan, Z.; Chen, M.; Goodman, S.; Gimpel, K.; Sharma, P.; and Soricut, R.
\newblock 2019.
\newblock Albert: A lite bert for self-supervised learning of language
  representations.
\newblock {\em arXiv preprint arXiv:1909.11942}.

\bibitem[\protect\citeauthoryear{Linzen, Dupoux, and
  Goldberg}{2016}]{linzen2016assessing}
Linzen, T.; Dupoux, E.; and Goldberg, Y.
\newblock 2016.
\newblock Assessing the ability of lstms to learn syntax-sensitive
  dependencies.
\newblock {\em Transactions of the Association for Computational Linguistics}
  4:521--535.

\bibitem[\protect\citeauthoryear{Liu \bgroup et al\mbox.\egroup
  }{2019}]{liu2019roberta}
Liu, Y.; Ott, M.; Goyal, N.; Du, J.; Joshi, M.; Chen, D.; Levy, O.; Lewis, M.;
  Zettlemoyer, L.; and Stoyanov, V.
\newblock 2019.
\newblock Roberta: A robustly optimized bert pretraining approach.
\newblock {\em arXiv preprint arXiv:1907.11692}.

\bibitem[\protect\citeauthoryear{Mao \bgroup et al\mbox.\egroup
  }{2022}]{mao2022biases}
Mao, R.; Liu, Q.; He, K.; Li, W.; and Cambria, E.
\newblock 2022.
\newblock The biases of pre-trained language models: An empirical study on
  prompt-based sentiment analysis and emotion detection.
\newblock {\em IEEE Transactions on Affective Computing}.

\bibitem[\protect\citeauthoryear{Marvin and Linzen}{2018}]{marvin2018targeted}
Marvin, R., and Linzen, T.
\newblock 2018.
\newblock Targeted syntactic evaluation of language models.
\newblock {\em arXiv preprint arXiv:1808.09031}.

\bibitem[\protect\citeauthoryear{McCoy, Pavlick, and
  Linzen}{2019}]{mccoy2019right}
McCoy, R.~T.; Pavlick, E.; and Linzen, T.
\newblock 2019.
\newblock Right for the wrong reasons: Diagnosing syntactic heuristics in
  natural language inference.
\newblock {\em arXiv preprint arXiv:1902.01007}.

\bibitem[\protect\citeauthoryear{Misra, Ettinger, and
  Rayz}{2020}]{misra2020exploring}
Misra, K.; Ettinger, A.; and Rayz, J.~T.
\newblock 2020.
\newblock Exploring bert's sensitivity to lexical cues using tests from
  semantic priming.
\newblock {\em arXiv preprint arXiv:2010.03010}.

\bibitem[\protect\citeauthoryear{Misra, Ettinger, and
  Rayz}{2021}]{misra2021language}
Misra, K.; Ettinger, A.; and Rayz, J.~T.
\newblock 2021.
\newblock Do language models learn typicality judgments from text?
\newblock {\em arXiv preprint arXiv:2105.02987}.

\bibitem[\protect\citeauthoryear{Misra, Rayz, and
  Ettinger}{2022}]{misra2022comps}
Misra, K.; Rayz, J.~T.; and Ettinger, A.
\newblock 2022.
\newblock Comps: Conceptual minimal pair sentences for testing property
  knowledge and inheritance in pre-trained language models.
\newblock {\em arXiv preprint arXiv:2210.01963}.

\bibitem[\protect\citeauthoryear{Nissim, van Noord, and van~der
  Goot}{2020}]{nissim2020fair}
Nissim, M.; van Noord, R.; and van~der Goot, R.
\newblock 2020.
\newblock Fair is better than sensational: Man is to doctor as woman is to
  doctor.
\newblock {\em Computational Linguistics} 46(2):487--497.

\bibitem[\protect\citeauthoryear{Pandia and Ettinger}{2021}]{pandia2021sorting}
Pandia, L., and Ettinger, A.
\newblock 2021.
\newblock Sorting through the noise: Testing robustness of information
  processing in pre-trained language models.
\newblock {\em arXiv preprint arXiv:2109.12393}.

\bibitem[\protect\citeauthoryear{Peters \bgroup et al\mbox.\egroup
  }{2018}]{peters2018dissecting}
Peters, M.~E.; Neumann, M.; Zettlemoyer, L.; and Yih, W.-t.
\newblock 2018.
\newblock Dissecting contextual word embeddings: Architecture and
  representation.
\newblock {\em arXiv preprint arXiv:1808.08949}.

\bibitem[\protect\citeauthoryear{Petroni \bgroup et al\mbox.\egroup
  }{2019}]{petroni2019language}
Petroni, F.; Rockt{\"a}schel, T.; Lewis, P.; Bakhtin, A.; Wu, Y.; Miller,
  A.~H.; and Riedel, S.
\newblock 2019.
\newblock Language models as knowledge bases?
\newblock {\em arXiv preprint arXiv:1909.01066}.

\bibitem[\protect\citeauthoryear{Quillian}{1967}]{quillian1967word}
Quillian, M.~R.
\newblock 1967.
\newblock Word concepts: A theory and simulation of some basic semantic
  capabilities.
\newblock {\em Behavioral science} 12(5):410--430.

\bibitem[\protect\citeauthoryear{Radford \bgroup et al\mbox.\egroup
  }{2019}]{radford2019language}
Radford, A.; Wu, J.; Child, R.; Luan, D.; Amodei, D.; and Sutskever, I.
\newblock 2019.
\newblock Language models are unsupervised multitask learners.

\bibitem[\protect\citeauthoryear{Rogers, Kovaleva, and
  Rumshisky}{2021}]{rogers2021primer}
Rogers, A.; Kovaleva, O.; and Rumshisky, A.
\newblock 2021.
\newblock A primer in bertology: What we know about how bert works.
\newblock {\em Transactions of the Association for Computational Linguistics}
  8:842--866.

\bibitem[\protect\citeauthoryear{Safavi and
  Koutra}{2021}]{safavi2021relational}
Safavi, T., and Koutra, D.
\newblock 2021.
\newblock Relational world knowledge representation in contextual language
  models: A review.
\newblock {\em arXiv preprint arXiv:2104.05837}.

\bibitem[\protect\citeauthoryear{Smith and Estes}{1978}]{smith1978theories}
Smith, E.~E., and Estes, W.~K.
\newblock 1978.
\newblock Theories of semantic memory.
\newblock {\em Handbook of learning and cognitive processes} 6:1--56.

\bibitem[\protect\citeauthoryear{Tenney \bgroup et al\mbox.\egroup
  }{2019}]{tenney2019you}
Tenney, I.; Xia, P.; Chen, B.; Wang, A.; Poliak, A.; McCoy, R.~T.; Kim, N.;
  Van~Durme, B.; Bowman, S.~R.; Das, D.; et~al.
\newblock 2019.
\newblock What do you learn from context? probing for sentence structure in
  contextualized word representations.
\newblock {\em arXiv preprint arXiv:1905.06316}.

\bibitem[\protect\citeauthoryear{Wilcox \bgroup et al\mbox.\egroup
  }{2018}]{wilcox2018rnn}
Wilcox, E.; Levy, R.; Morita, T.; and Futrell, R.
\newblock 2018.
\newblock What do rnn language models learn about filler-gap dependencies?
\newblock {\em arXiv preprint arXiv:1809.00042}.

\bibitem[\protect\citeauthoryear{Zhao \bgroup et al\mbox.\egroup
  }{2018}]{zhao2018gender}
Zhao, J.; Wang, T.; Yatskar, M.; Ordonez, V.; and Chang, K.-W.
\newblock 2018.
\newblock Gender bias in coreference resolution: Evaluation and debiasing
  methods.
\newblock {\em arXiv preprint arXiv:1804.06876}.

\end{thebibliography}
\end{document}